\documentclass[runningheads]{llncs}

\usepackage{eccv}

\usepackage{eccvabbrv}

\usepackage{graphicx}
\usepackage{booktabs}
\usepackage{array}
\usepackage{colortbl}
\usepackage{multirow}
\usepackage{adjustbox}
\usepackage{float}

\usepackage[accsupp]{axessibility}  

\usepackage{hyperref}

\usepackage{orcidlink}

\definecolor{col3view}{RGB}{220,235,255}
\definecolor{col6view}{RGB}{230,255,230} 
\definecolor{col9view}{RGB}{255,240,225}
\newcolumntype{B}{>{\columncolor{col3view}}c}
\newcolumntype{G}{>{\columncolor{col6view}}c}
\newcolumntype{P}{>{\columncolor{col9view}}c}

\definecolor{bestbg}{RGB}{243,185,184}
\definecolor{secbg}{RGB}{255,225,196}
\definecolor{thirdbg}{RGB}{255,251,206}
\newcommand{\best}[1]{\cellcolor{bestbg}#1}
\newcommand{\secondbest}[1]{\cellcolor{secbg}#1}
\newcommand{\thirdbest}[1]{\cellcolor{thirdbg}#1}

\newcommand{\tabstyle}{\setlength{\tabcolsep}{2pt}}

\newcommand{\bo}[1]{\best{#1}}
\let\bs\secondbest
\let\bt\thirdbest

\begin{document}

\title{Improving Sparse-View 3DGS Generalization\\via Flat Minima Optimization} 


\author{Kangmin Seo\orcidlink{0009-0008-0288-4889} \and
Sangeek Hyun\orcidlink{0000-0002-4050-6896} \and
MinKyu Lee\orcidlink{0000-0003-3821-8745} \and
Jae-Pil Heo\thanks{Corresponding author}\orcidlink{0000-0001-9684-7641}}

\authorrunning{K. Seo et al.}

\institute{Sungkyunkwan University, South Korea}

\maketitle

\begin{abstract}
Recent advances in neural rendering have established 3D Gaussian Splatting (3DGS) as a highly efficient representation for novel view synthesis, enabling fast training and real-time rendering with strong fidelity. However, when supervision is limited to sparse input views, 3DGS tends to overfit to the observed images and generalize poorly to unseen viewpoints. We address this challenge from the perspective of flat minima (FM) optimization, which seeks solutions that remain stable under small parameter perturbations. Viewing Gaussian parameters as trainable weights, we adapt FM principles to the geometric and dynamic nature of 3DGS with a lightweight training framework. Our method regularizes optimization with controlled Gaussian perturbations that account for each Gaussian's anisotropy and the training progress, preserving fine details while improving robustness to sparse-view overfitting. To further stabilize this flat minima optimization process, we introduce periodic reinitialization, which temporarily returns non-positional parameters to their initial states for a short window. Together, these techniques integrate seamlessly into existing 3DGS pipelines without architectural changes. Experiments on LLFF and Mip-NeRF360 datasets demonstrate improved quantitative metrics and perceptual quality under sparse-view supervision, producing reconstructions that are sharper, more stable, and better generalized to novel viewpoints.


\end{abstract}

\section{Introduction}

Recent advances in neural rendering have been driven by representative methods such as Neural Radiance Fields (NeRF)~\cite{nerf} and, more recently, 3D Gaussian Splatting (3DGS)~\cite{3dgs}. While these approaches have achieved remarkable progress in novel view synthesis (NVS), sparse-view scenarios, where only a few input images are available, remain highly challenging. In such settings, models are prone to overfitting to the input views, leading to poor generalization to novel viewpoints.

In this paper, we aim to tackle this through the lens of flat minima (FM) optimization. Originally developed in the context of neural networks, FM optimization improves generalization by encouraging solutions that reside in flatter regions of the loss landscape, where small parameter perturbations do not significantly increase the loss~\cite{flat, sharpminima, onlargebatch}. Inspired by this principle, we reinterpret 3DGS as a supervised learning system where camera poses serve as inputs, rendered images as outputs, and Gaussian parameters play the role of learnable weights. Within this formulation, overfitting in 3DGS corresponds to sharp-minima solutions that are highly sensitive to parameter shifts; resembling the failure modes addressed by FM theory.

Motivated by these insights, we propose to introduce flat minima optimization into the sparse-view 3DGS setting. Specifically, we apply stochastic perturbations to Gaussian parameters during training, following the spirit of prior FM techniques that encourage robustness through parameter noise injection~\cite{rwp, revisitingrwp}. At each iteration, random noise is added to Gaussian parameters, and the model is trained to minimize reconstruction loss under these perturbations, guiding it toward flatter minima and more generalizable solutions.

Beyond this baseline application of flat minima optimization, we further tailor the approach to the specific characteristics of 3DGS. While directly adding perturbations can improve generalization, naïve strategies often suppress fine-grained geometry, leading to underfitting of local structures. To address this, we adapt FM optimization to better align with the geometric properties and training dynamics of 3DGS.

First, we introduce a Scale-Adaptive Perturbation (SAP) strategy that takes into account the scale and anisotropic shape of each Gaussian. Perturbation noise is sampled in proportion to the Gaussian’s spatial extent, applying larger perturbations along longer axes and smaller ones along shorter axes. This ensures that perturbations remain meaningful across primitives of varying sizes, preventing destabilization for small Gaussians while still enforcing robustness for larger ones. In particular, this design helps preserve fine-grained scene details while maintaining the regularization benefits of flat minima optimization.

We adopt a stochastic perturbation scheme inspired by prior FM studies ~\cite{revisitingrwp}, which combine losses from perturbed and unperturbed models to encourage convergence to flatter solutions. Building on this idea, we instead apply SAP in a stochastic manner. That is, for each iteration, each Gaussian is probabilistically perturbed or left unchanged. This avoids the need to render both perturbed and unperturbed models separately, while keeping the training objective simple and lightweight.

In addition, we complement positional perturbations with periodic parameter reinitialization. Scale, rotation, and higher-order SH coefficients are periodically returned to their initial state, following the same initialization used when constructing Gaussians from SfM points, and kept unchanged for a short window. Opacity is reset at each interval following the standard mechanism in 3DGS~\cite{3dgs}, while positions remain unaffected. This acts as an additional regularizer that mitigates overfitting and stabilizes training under sparse-view supervision.

Overall, our framework is lightweight, requires no architectural modifications, and integrates seamlessly into existing pipelines. Experiments demonstrate that it consistently improves both quantitative metrics and perceptual quality, producing reconstructions that are sharper, more stable, and better generalized to novel viewpoints.

\section{Related Work}
3D Gaussian Splatting has attracted substantial attention for novel view synthesis, but sparse-view settings remain challenging because limited input views make the optimization underconstrained, often leading to degraded novel view quality. A broad set of recent methods improve robustness by introducing additional geometric cues, imposing constraints on geometric structure, or regularizing primitive interactions.

A common direction is to leverage geometric priors, especially depth-related supervision or normalization, to stabilize geometry under sparse observations. Representative examples include depth-aided training and normalization strategies that reduce geometric ambiguity and improve reconstruction and rendering in sparse-view settings \cite{fsgs,dngaussian,drgs,nexusgs,d2gs,ugod}. Related methods also refine geometry through additional constraints or alternative training schedules, such as co-regularization and alternating densification behaviors that affect how primitives are created, split, and pruned during training \cite{corgs,adgs}. Other works focus on sparse-view reconstruction with geometry-prioritized formulations, including approaches that emphasize surface consistency or surface reconstruction objectives \cite{sparse2dgs,sparsegs}.

Another line of work improves sparse-view generalization through stochastic regularization of the primitive set. DropGaussian and DropoutGS randomly remove Gaussians during training to discourage brittle reliance on a small subset of primitives and to promote more distributed explanations \cite{dropgaussian,dropoutgs}. The co-adaptation analysis further formalizes this phenomenon by quantifying inter-primitive entanglement and alleviating it with Gaussian dropout and opacity noise injection \cite{co-adaptation}. These methods primarily regularize the representation structurally by changing which primitives contribute and how primitives interact.

Self-ensembling and consistency-based approaches also improve robustness. SE-GS introduces uncertainty-aware perturbations and self-ensembling objectives to stabilize sparse-view optimization \cite{segs}, while SCGaussian uses matching priors to enforce structure consistency \cite{scgaussian}.

A recent method leverages a video diffusion prior with scene-grounding guidance to synthesize additional observations for sparse inputs, which are then used to optimize 3DGS \cite{tamingvideo}. Difix3D+ enhances reconstruction and novel-view synthesis via single-step diffusion models as a learned prior that helps correct artifacts under sparse-view conditions \cite{difix3dplus}. MAtCha Gaussians proposes an atlas-of-charts surface representation rendered with 2D Gaussian surfels, initialized with monocular depth and refined for photorealistic rendering from sparse views \cite{matcha}. These pipelines can be highly effective, but they typically rely on external priors and additional components beyond standard 3DGS optimization.

Finally, structural regularization has been studied to reduce degenerate Gaussian configurations. Prior methods regularize geometry, primitive structure, or inter-primitive interactions by introducing geometric cues or constraints, penalizing undesirable covariance shapes, or changing how Gaussian primitives contribute and interact \cite{fsgs,dngaussian,drgs,nexusgs,d2gs,ugod,erank,dropgaussian,dropoutgs,co-adaptation,segs}.
\section{Method}

\subsection{Preliminaries}

\textbf{3D Gaussian Splatting~(3DGS)} represents a scene using a set of anisotropic Gaussians, each defined by learnable parameters including position $x_i \in \mathbb{R}^3$, scale $s_i$, rotation $r_i$, color $c_i$, and opacity $o_i$, where $i$ indexes the $i$-th Gaussian.
These Gaussians are projected and composited through differentiable rasterization to render photorealistic images from arbitrary viewpoints.

Given a set of training images and corresponding camera parameters, the Gaussian parameters are optimized by minimizing a photometric reconstruction loss:

\begin{equation}
\theta \leftarrow \theta - \eta \cdot \nabla \mathcal{L}(\theta),
\end{equation}

where $\theta$ encompasses all learnable parameters and $\mathcal{L}(\theta)$ denotes the rendering loss, typically computed as a combination of L1 and SSIM losses between the rendered and corresponding ground-truth images, and $\eta$ is the learning rate.

\begin{figure*}[tb]
    \begin{center}
    \includegraphics[width=\linewidth]{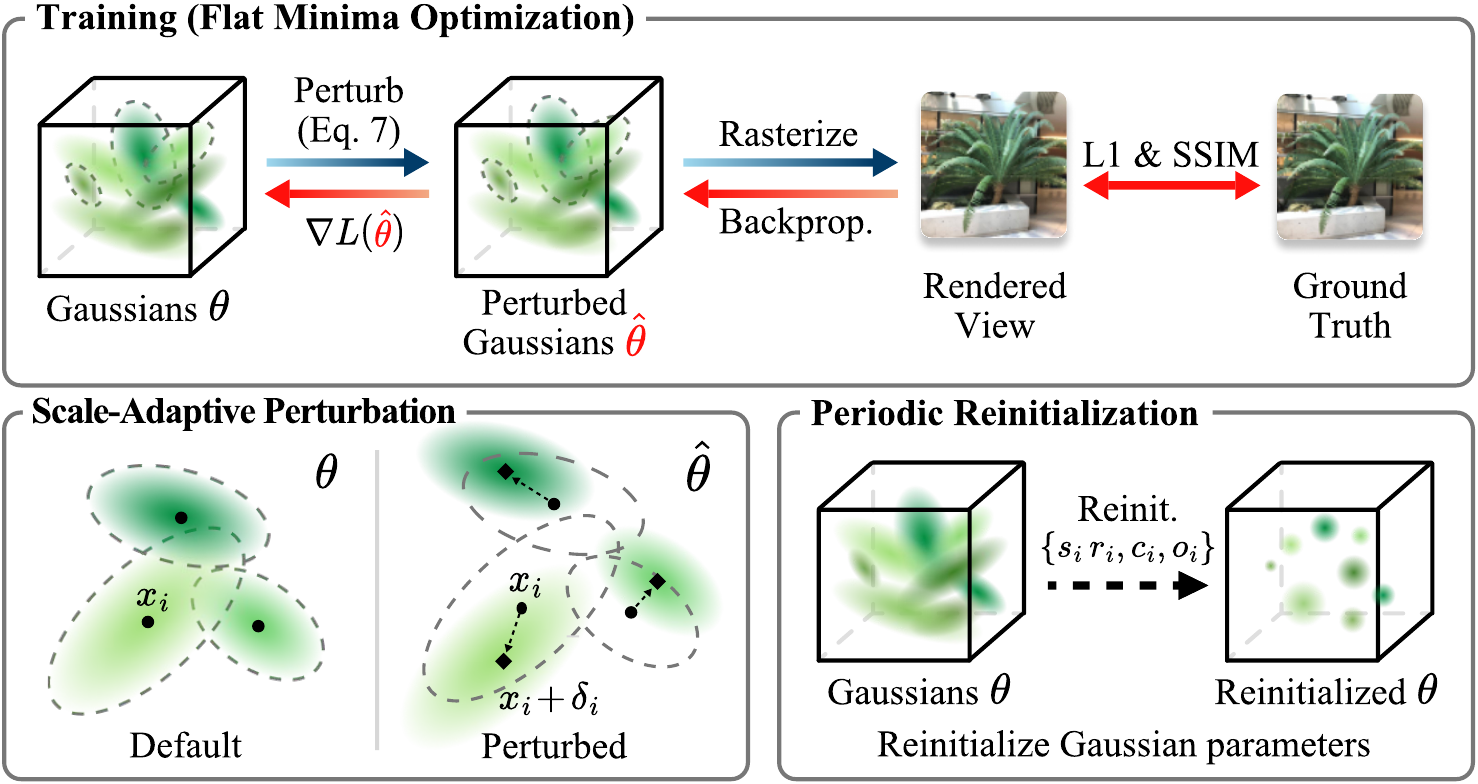}
    \end{center}
    \caption{
        Overview of our flat minima optimization framework for sparse-view 3D Gaussian Splatting (3DGS). Our approach enhances generalization by introducing perturbations to Gaussian positions during training. To account for the geometric structure, we scale perturbations according to the anisotropic extent of each Gaussian and apply them stochastically at the level of individual Gaussians. In addition, we periodically reinitialize non-positional parameters by temporarily returning scale, rotation, and higher-order SH coefficients to their initial state, following the same initialization method used when constructing Gaussians from SfM points. These parameters are kept unchanged for a short window, while opacity is reset following the standard mechanism in 3DGS.
    }
    \label{fig:main_figure}
\end{figure*}

\begin{figure*}[tb]
    \centering
    \includegraphics[width=\linewidth]{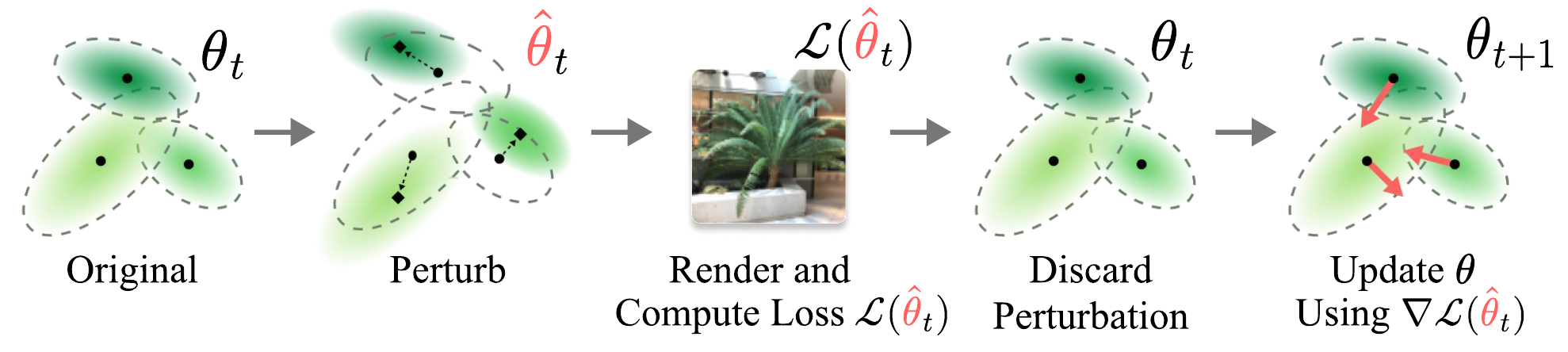}
    \caption{
        Optimization procedure of our method. At iteration $t$, we temporarily perturb the Gaussian model to obtain $\hat{\theta}_t$, render it, and compute the standard 3DGS reconstruction loss $\mathcal{L}(\hat{\theta}_t)$ without introducing an additional loss term or auxiliary objective. The perturbation is then discarded, while $\nabla \mathcal{L}(\hat{\theta}_t)$ updates the original model from $\theta_t$ to $\theta_{t+1}$.
    }
    \label{fig:optimization_procedure}
\end{figure*}

\textbf{Flat Minima~(FM) Optimization} aims to improve generalization by encouraging the model to converge to regions in the loss landscape where small perturbations do not significantly increase the loss. Here, the representative frameworks are adversarial and stochastic perturbation-based optimization. Sharpness-Aware Minimization (SAM)~\cite{sharpness-aware} is a representative of the former method, raising the adversarial perturbation scheme. It minimizes the worst-case loss within a small neighborhood around the parameters as following:

\begin{equation}
\mathcal{L}_{\text{SAM}}(\theta) = \min_{\theta} \max_{\|\epsilon\|_2 \leq \rho} \mathcal{L}(\theta + \epsilon).
\end{equation}

This min-max formulation aggressively biases optimization toward flatter regions of the loss landscape, but requires an additional gradient computation, which significantly increases the training cost. Meanwhile, Random Weight Perturbation (RWP)~\cite{lpfsgd,rwp} provides a more efficient alternative by replacing the worst-case loss with an expected loss over random perturbations:

\begin{equation} 
\mathcal{L}_{\text{RWP}}(\theta) = \mathbb{E}_{\epsilon \sim \mathcal{N}(0, \sigma^2 I)} \left[ \mathcal{L}(\theta + \epsilon) \right].
\end{equation}

By averaging losses across randomly perturbed weight configurations, RWP effectively smooths the loss landscape without additional backpropagation. Later works~\cite{revisitingrwp} further refined this approach by mixing perturbed and unperturbed losses and adapting noise magnitudes, achieving stronger performance with minimal cost.

In this work, we build upon these insights and adapt flat minima optimization to the context of 3D Gaussian Splatting, as summarized in \cref{fig:main_figure}.

\subsection{Flat Minima Optimization in 3DGS}
We begin by formulating flat minima (FM) optimization in the context of 3D Gaussian Splatting (3DGS). While conventional FM approaches in feed-forward neural networks inject random perturbations into all weights to encourage flatter solutions, directly adopting this strategy in 3DGS may overlook its unique geometric structure. Instead, we reinterpret FM optimization by perturbing the 3D position coordinates $\mathbf{x}_i$ of Gaussians during training. From a geometric perspective, overfitting in sparse-view 3DGS manifests as sharp minima that are highly sensitive to positional shifts, which in turn causes vulnerability in novel view rendering. By introducing position-level perturbations, we guide the optimization toward flatter minima where Gaussian positions are more robust under spatial variation. Accordingly, we extend this idea with perturbation and reinitialization techniques that account for the geometric properties and dynamic behavior of 3DGS.

\paragraph{Scale-Adaptive Perturbation~(SAP).}
Prior works on random perturbation strategies~\cite{lpfsgd, revisitingrwp} have demonstrated the effectiveness of stochastic perturbations in improving generalization. To adapt this principle to the geometric nature of 3DGS, we introduce a Scale-Adaptive Perturbation (SAP) scheme that tailors the perturbation magnitude to the anisotropic scale of each Gaussian. By aligning perturbations with the spatial extent of each primitive, SAP effectively balances robustness and detail preservation in 3D reconstruction.

Formally, the perturbed parameters $\hat{\theta}$ are defined as
\begin{equation}
\begin{aligned}
\mathbf{x}_i' = \mathbf{x}_i + \delta_i,
\quad
\delta_i \sim \mathcal{N}\!\bigl(0, \gamma^2 \mathbf{\Sigma}_i\bigr),\\
\mathbf{\Sigma}_i = \mathbf{R}_i\, \mathbf{S}_i \mathbf{S}_i^{\top} \mathbf{R}_i^{\top},
\quad
\hat{\theta} = \{\mathbf{x}_i', s_i, r_i, c_i, o_i\},
\end{aligned}
\end{equation}

where $\mathbf{x}_i$ denotes the 3D position of the $i$-th Gaussian, 
$s_i$ is its anisotropic scale, 
$r_i$ its rotation, 
$c_i$ its color, 
and $o_i$ its opacity. 
Here, $\mathbf{R}_i$ is the rotation matrix constructed from $r_i$, 
and $\mathbf{S}_i = \mathrm{diag}(s_i)$ is the scaling matrix whose diagonal elements correspond to the per-axis scales of the Gaussian. As illustrated in \cref{fig:optimization_procedure}, gradients are then computed at the perturbed parameters $\hat{\theta}$, and the update at iteration $t$ is performed accordingly:
\begin{equation}
\theta_{t+1} \leftarrow \theta_t - \eta \cdot \nabla \mathcal{L}(\hat{\theta}_t).
\end{equation}

Unlike 3DGS-MCMC~\cite{3dgs-mcmc}, which injects noise into the parameter update ($\theta_{t+1} = \theta_t - \eta \nabla \mathcal{L}(\theta_t) + \epsilon_t$) so that stochasticity accumulates along the trajectory, our method perturbs only the loss evaluation while keeping the parameters clean, locally smoothing the loss landscape toward flatter minima.

By modulating perturbations according to the anisotropic scale of each Gaussian, SAP enforces stronger regularization on larger or elongated primitives while applying finer noise to smaller ones. This design preserves high-frequency details in compact Gaussians while still promoting robustness across the broader scene geometry.

\paragraph{Stochastic Application of SAP.}
To further adapt flat minima optimization to 3DGS, we introduce stochastic perturbations, where each Gaussian is probabilistically perturbed or left unchanged. This design is motivated by recent Random Weight Perturbation strategies~\cite{revisitingrwp}, where perturbed and unperturbed objectives are mixed to encourage flatter solutions. Instead of explicitly rendering two models and combining their losses, we achieve a similar effect more efficiently by perturbing each Gaussian with probability $p$ at every iteration. In practice, this avoids the need to perform two forward passes, keeping the training objective simple and lightweight.

Formally, for each Gaussian $i$, the perturbed position $\mathbf{x}_i'$ is defined as
\begin{equation}
\mathbf{x}_i' =
\begin{cases}
\mathbf{x}_i + \delta_i, & \text{with probability } p, \\
\mathbf{x}_i, & \text{with probability } 1-p,
\end{cases}
\end{equation}

Applying perturbations in this localized, probabilistic manner offers an additional advantage over global perturbation strategies. In prior FM-inspired methods, where perturbed losses are always incorporated into optimization, we empirically observe a tendency to suppress fine-grained geometry, leading to underfitting of local structures. Similar observations can also be found in our ablation results (see \cref{tab:ablation_loss_mix}), where the loss-mixing strategy shows degradation in rendering fidelity. By contrast, our per-Gaussian stochastic formulation limits the extent to which perturbations dominate the training signal, allowing the model to retain delicate scene details while still benefiting from the regularization effect. This not only reduces overfitting but also enhances the stability of optimization without incurring additional computational overhead.

\paragraph{Perturbation Magnitude Scheduling.}
We linearly scale the perturbation magnitude from $0$ at the beginning to a target value at the final iteration. This gradual increase avoids excessive noise in the early stage, while ensuring that the desired perturbation strength is applied once the model reaches a more stable regime. Empirically, we find that applying strong perturbations too early, before Gaussians have sufficiently captured the scene structure, tends to degrade performance, motivating this progressive schedule.

Formally, the stochastic perturbation is reformulated to incorporate this scheduling scheme, following the linear annealing strategy used in prior work such as~\cite{dropgaussian}:
\begin{equation}
\mathbf{x}_i' =
\begin{cases}
\mathbf{x}_i + \alpha(t)\,\delta_i, & \text{with probability } p, \\
\mathbf{x}_i, & \text{with probability } 1-p,
\end{cases}
\quad\text{where}\quad
\alpha(t) = \tfrac{t}{T},
\end{equation}
where $t$ is the current iteration, $T$ is the total number of iterations, and $\alpha(t)$ linearly increases the perturbation magnitude from $0$ to $1$ over training.

\paragraph{Periodic Gaussian Reinitialization.}
In addition to position-level perturbations, which serve as our main mechanism for flat minima optimization, we periodically reinitialize a subset of Gaussian parameters during training. Specifically, at fixed intervals we temporarily return scale, rotation, and higher-order SH coefficients to their initial state, following the same initialization procedure used when constructing Gaussians from a Structure-from-Motion (SfM)~\cite{sfm} point cloud, and keep them unchanged for $W$ iterations. Opacity is reset at each interval following the standard reset mechanism in 3DGS~\cite{3dgs}. Positions and the total number of Gaussians remain unaffected. This design complements position perturbations by providing an additional regularization effect that mitigates overfitting under sparse-view supervision. In particular, this return-to-initialization step reduces the effective degrees of freedom of non-positional parameters during the window, which stabilizes optimization. By periodically applying this process during training, we enhance stability and further guide optimization toward flatter minima.

\section{Experiments}

\begin{table*}[tb]
  \centering
  \caption{Quantitative comparison on the LLFF dataset under 3-view, 6-view, and 9-view settings. Our method achieves consistently strong performance across all metrics, outperforming or matching prior baselines in PSNR, SSIM, and LPIPS. The results demonstrate the effectiveness of our flat-minima optimization framework in improving both photometric accuracy and perceptual quality.}
  \footnotesize
  \tabstyle
  \renewcommand{\arraystretch}{1.1}
  \begin{adjustbox}{max width=\linewidth}
      \begin{tabular}{l ccc ccc ccc}
        \toprule
        \multirow{2}{*}{Method} &
          \multicolumn{3}{c}{\textbf{3-view}} &
          \multicolumn{3}{c}{\textbf{6-view}} &
          \multicolumn{3}{c}{\textbf{9-view}} \\
        \cmidrule(lr){2-4}\cmidrule(lr){5-7}\cmidrule(lr){8-10}
          & PSNR\,$\uparrow$ & SSIM\,$\uparrow$ & LPIPS\,$\downarrow$ &
            PSNR\,$\uparrow$ & SSIM\,$\uparrow$ & LPIPS\,$\downarrow$ &
            PSNR\,$\uparrow$ & SSIM\,$\uparrow$ & LPIPS\,$\downarrow$ \\
        \midrule
        3DGS          & 19.22 & 0.649 & 0.229 & 23.80 & 0.814 & 0.125 & 25.44 & 0.860 & 0.096 \\
        DNGaussian    & 19.12 & 0.591 & 0.294 & 22.18 & 0.755 & 0.198 & 23.17 & 0.788 & 0.180 \\
        FSGS          & 20.43 & 0.682 & 0.248 & 24.09 & 0.823 & 0.145 & 25.31 & 0.860 & 0.122 \\
        CoR-GS        & \thirdbest{20.45}  & \thirdbest{0.712} & \secondbest{0.196} &
                        \thirdbest{24.49} & \secondbest{0.837} & \secondbest{0.115} &
                        \thirdbest{26.06}  & \secondbest{0.874} & \secondbest{0.089}  \\
        DropGaussian  & \secondbest{20.76} & \secondbest{0.713} & \thirdbest{0.200} &
                        \secondbest{24.74} & \secondbest{0.837} & \thirdbest{0.117} &
                        \secondbest{26.21} & \secondbest{0.874} & \best{0.088} \\
        \midrule
        \textbf{Ours} & \best{20.88} & \best{0.731} & \best{0.184} &
                        \best{24.76} & \best{0.840} & \best{0.114} &
                        \best{26.23} & \best{0.875} & \best{0.088} \\
        \bottomrule
      \end{tabular}
      
      \label{tab: quantitative llff}
    \end{adjustbox}
\end{table*}

\begin{table}[tb]
    \footnotesize
    \centering
    \caption{Quantitative comparison on MipNeRF360 dataset under 12-view and 24-view settings.}
    \label{tab:quantitative_mip360}
    \renewcommand{\arraystretch}{0.96}
    \setlength{\tabcolsep}{4.5pt}
    \begin{adjustbox}{max width=\linewidth}
      \begin{tabular}{l ccc ccc}
        \toprule
        \multirow{2}{*}{Method} &
          \multicolumn{3}{c}{\textbf{12-view}} &
          \multicolumn{3}{c}{\textbf{24-view}} \\[0.5pt]
        \cmidrule(lr){2-4}\cmidrule(lr){5-7}
          & PSNR\,$\uparrow$ & SSIM\,$\uparrow$ & LPIPS\,$\downarrow$ &
            PSNR\,$\uparrow$ & SSIM\,$\uparrow$ & LPIPS\,$\downarrow$ \\
        \midrule
        3DGS          & 18.52 & 0.523 & \thirdbest{0.415} & 22.80 & 0.708 & 0.276 \\
        FSGS          & 18.80 & 0.531 & 0.418 & \thirdbest{23.70} & \thirdbest{0.745} & \thirdbest{0.230} \\
        CoR-GS        & \secondbest{19.52} & \thirdbest{0.558} & 0.418 &
                        23.39 & 0.727 & 0.271 \\
        DropGaussian  & \best{19.74} & \secondbest{0.577} & \secondbest{0.364} &
                        \secondbest{24.13} & \secondbest{0.762} & \secondbest{0.225} \\
        \midrule
        \textbf{Ours} & \thirdbest{19.50} & \best{0.584} & \best{0.348} &
                        \best{24.19} & \best{0.771} & \best{0.219} \\
        \bottomrule
      \end{tabular}
    \end{adjustbox}
\end{table}

\begin{figure*}[tb]
    \centering
    \setlength{\tabcolsep}{2pt}
    \newcommand{\quallabel}[1]{\makebox[0.16\textwidth][c]{#1}}
    \resizebox{\linewidth}{!}{
    \begin{tabular}{@{}cccccc@{}}
        \includegraphics[width=0.16\textwidth]{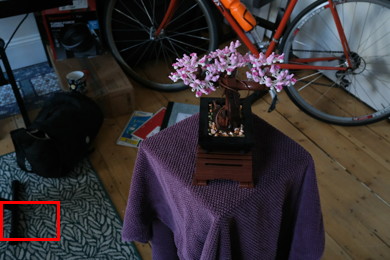} & \includegraphics[width=0.16\textwidth]{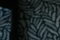} & \includegraphics[width=0.16\textwidth]{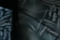} & \includegraphics[width=0.16\textwidth]{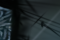} & \includegraphics[width=0.16\textwidth]{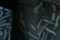} & \includegraphics[width=0.16\textwidth]{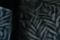} \\
        \includegraphics[width=0.16\textwidth]{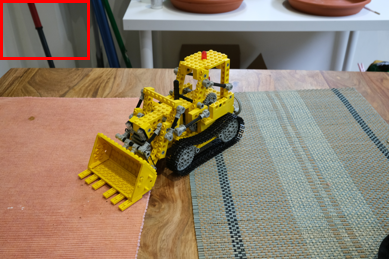} & \includegraphics[width=0.16\textwidth]{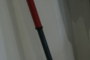} & \includegraphics[width=0.16\textwidth]{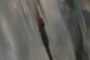} & \includegraphics[width=0.16\textwidth]{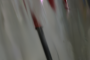} & \includegraphics[width=0.16\textwidth]{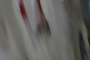} & \includegraphics[width=0.16\textwidth]{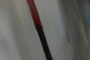} \\
        \includegraphics[width=0.16\textwidth]{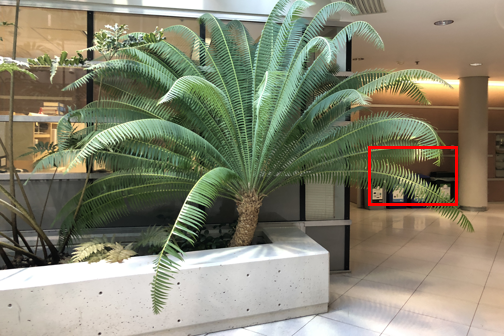} & \includegraphics[width=0.16\textwidth]{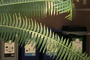} & \includegraphics[width=0.16\textwidth]{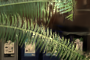} & \includegraphics[width=0.16\textwidth]{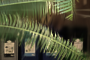} & \includegraphics[width=0.16\textwidth]{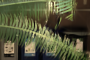} & \includegraphics[width=0.16\textwidth]{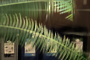} \\
        \includegraphics[width=0.16\textwidth]{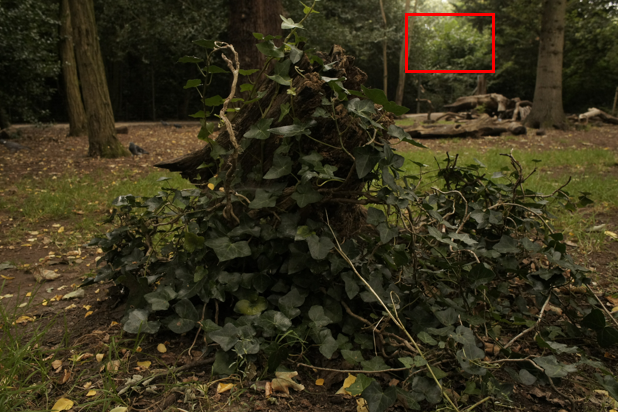} & \includegraphics[width=0.16\textwidth]{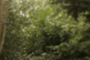} & \includegraphics[width=0.16\textwidth]{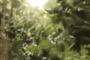} & \includegraphics[width=0.16\textwidth]{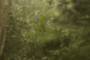} & \includegraphics[width=0.16\textwidth]{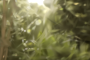} & \includegraphics[width=0.16\textwidth]{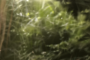} \\
        \includegraphics[width=0.16\textwidth]{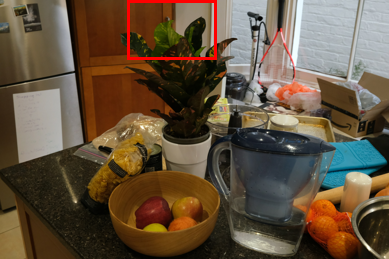} & \includegraphics[width=0.16\textwidth]{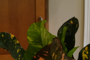} & \includegraphics[width=0.16\textwidth]{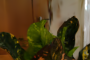} & \includegraphics[width=0.16\textwidth]{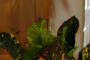} & \includegraphics[width=0.16\textwidth]{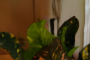} & \includegraphics[width=0.16\textwidth]{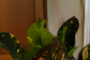} \\
        \quallabel{GT full-view} &
        \quallabel{GT crop} &
        \quallabel{3DGS} &
        \quallabel{CoR-GS} &
        \quallabel{DropGaussian} &
        \quallabel{Ours} \\
    \end{tabular}
    }
    \caption{Qualitative comparison in LLFF and Mip-NeRF360 datasets. Compared to prior methods, our approach reconstructs sharper details and more geometrically consistent structures, especially in under-constrained regions such as planar surfaces and occlusion boundaries. While baselines often suffer from spatial misalignment or blurring, our method produces visually coherent and high-fidelity results, demonstrating improved generalization to novel viewpoints.}

    \label{fig:main_qualitative}
\end{figure*}

\paragraph{Experimental Setup.}
We base our implementation on the original 3DGS framework~\cite{3dgs}, while following most hyperparameters from DropGaussian~\cite{dropgaussian}. We set the perturbation probability to $p_{\max}=0.3$, perturbation coefficient $\gamma=2$, and reinitialize scale, rotation, and higher-order SH coefficients every $1{,}000$ iterations, returning them to their SfM-based initial state for $W=100$ iterations; opacity follows the standard reset in~\cite{3dgs} at each interval. To ensure stability, perturbed positions are clamped so that the displacement does not exceed the corresponding Gaussian’s scale. Experiments were conducted on NVIDIA RTX TITAN and A6000 GPU.

\paragraph{Datasets and Evaluation Metrics.}

We evaluate our method on two widely used benchmarks for novel view synthesis: LLFF~\cite{llff} and Mip-NeRF360~\cite{mipnerf360}, both downsampled by a factor of 8.
For LLFF, we use 3, 6, and 9 input views, and for Mip-NeRF360, we use 12 and 24 views, following prior works. The reconstruction performance is evaluated by three commonly adopted image quality metrics: peak signal-to-noise ratio (PSNR), structural similarity index (SSIM)~\cite{ssim}, and learned perceptual image patch similarity (LPIPS)~\cite{lpips}.
Together, these metrics provide a comprehensive evaluation of both photometric accuracy and perceptual quality of novel view synthesis.

\paragraph{Baselines.}
We compare our method with recent approaches for sparse-view novel view synthesis, including 3DGS~\cite{3dgs}, CoR-GS~\cite{corgs}, and DropGaussian~\cite{dropgaussian}.
We also include DNGaussian~\cite{dngaussian} and FSGS~\cite{fsgs}, which incorporate additional geometric supervision using depth priors.

\subsection{Quantitative Evaluation}
\cref{tab: quantitative llff} and \cref{tab:quantitative_mip360} summarize the quantitative results across all sparse-view benchmarks. Our method outperforms the compared prior approaches across datasets and view settings. Notably, improvements are observed in PSNR, SSIM and perceptual quality as measured by LPIPS. These results demonstrate the effectiveness of our flat minima framework in enhancing both reconstruction fidelity and generalization under sparse-view supervision.

\subsection{Qualitative Evaluation}
\cref{fig:main_qualitative} presents qualitative comparisons across various sparse-view scenarios. Compared to existing baselines, our method consistently generates sharper and more structurally coherent renderings, particularly under challenging geometric conditions.

Baseline methods often suffer from noticeable blurring in background regions where supervision is scarce. These methods also tend to exhibit structural inconsistencies, such as spatial misalignment or distorted geometry. For instance, CoR-GS and DropGaussian occasionally fail to maintain consistent geometry, leading to spatial shifts in novel viewpoints, or inaccurately rendering planar surfaces and straight lines (\eg, second row). Such artifacts indicate insufficient regularization or overfitting to the limited input views.

In contrast, our method produces clean and sharper reconstructions across all scenes. The recovered geometry remains consistent under significant viewpoint changes, preserving fine-grained details such as object boundaries and linear structures. These improvements highlight the effectiveness of our flat minima framework in promoting generalizable geometry, and its robustness under sparse-view supervision.

\subsection{Perturbation Robustness Comparison}
\begin{figure*}[tb]
    \centering
    \begin{subfigure}[t]{0.48\textwidth}
        \centering
        \includegraphics[width=\linewidth, trim=0 0 0 0, clip]{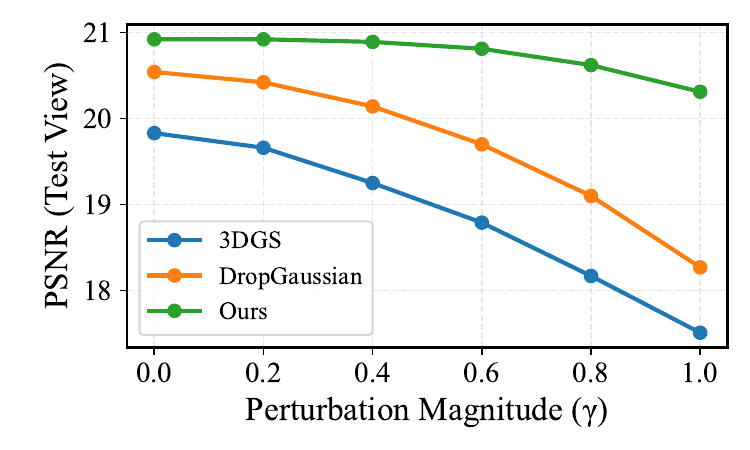}
    \end{subfigure}
    \begin{subfigure}[t]{0.48\textwidth}
        \centering
        \includegraphics[width=\linewidth, trim=0 0 0 0, clip]{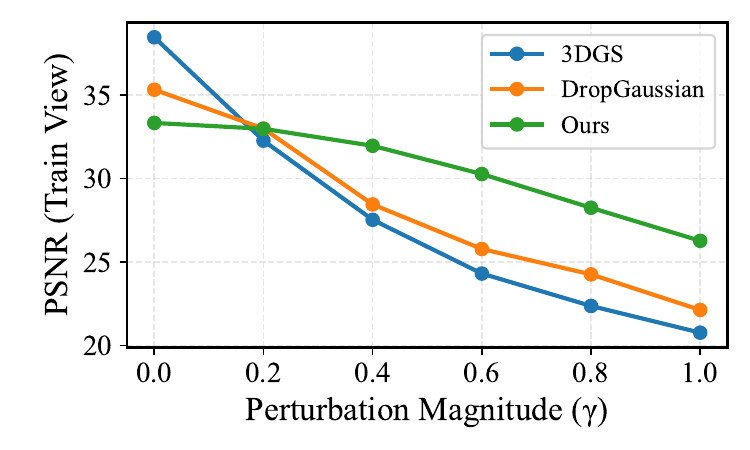}
    \end{subfigure}
    \caption{Perturbation robustness analysis on LLFF dataset under 3-view setting. Our method shows smaller performance degradation across perturbation magnitudes compared to 3DGS~\cite{3dgs} and DropGaussian~\cite{dropgaussian}.}
    \label{fig:flatness_llff3_test}
\end{figure*}

In line with prior works that evaluate the sharpness of minima through a model's
sensitivity to parameter perturbations
\cite{onlargebatch, sharpminima, sharpness-aware},
we examine how the reconstruction quality of a trained 3DGS model changes when its
Gaussian positions are perturbed after optimization.
We apply position perturbations using our Scale-Adaptive Perturbation (SAP)
with magnitudes
$\gamma \in \{0.0, 0.2, 0.4, 0.6, 0.8, 1.0\}$
and evaluate the resulting models on both training and test views.

As shown in \cref{fig:flatness_llff3_test},
our method shows substantially smaller degradation than both
3DGS~\cite{3dgs} and DropGaussian~\cite{dropgaussian} across all perturbation
magnitudes.
On test views, DropGaussian nearly matches the degradation trend of vanilla
3DGS, while our method shows substantially smaller drops at every scale.
The difference is more evident in training views as vanilla 3DGS and
DropGaussian exhibit larger degradation, indicating a stronger tendency
to overfit to training images compared to our method. This suggests that our optimization strategy leads to flatter and more robust minima than existing 3DGS baselines.

\begin{table}[tb]
    \footnotesize
    \centering
    \caption{Ablation studies on the LLFF dataset under the 3-view setting. $\dagger$ indicates our default setting.}
    \label{tab:ablation_noisetype}
    \label{tab:ablation_noiseparam}
    \label{tab:ablation_combined}
    \label{tab:ablation_loss_mix}
    \label{tab:ablation_schedule}
    \label{tab:ablation_reinit}
    \renewcommand{\arraystretch}{1.0}
    \setlength{\tabcolsep}{8pt}
    \setlength{\aboverulesep}{0.35ex}
    \setlength{\belowrulesep}{0.35ex}
    \begin{tabular}{l ccc}
      \toprule
      Method & PSNR & SSIM & LPIPS \\
      \midrule
      \multicolumn{4}{l}{\textit{Noise distribution}} \\
      \midrule
      Anisotropic$^\dagger$ & \bo{20.88} & \bo{0.731} & \bo{0.184} \\
      Isotropic (max)       & \bs{20.73} & \bs{0.725} & \bt{0.188} \\
      Isotropic (mean)      & \bt{20.67} & \bt{0.724} & \bs{0.185} \\
      Isotropic (fixed)     & 20.54 & 0.714 & \bt{0.188} \\
      \midrule
      \multicolumn{4}{l}{\textit{Perturbed parameter}} \\
      \midrule
      Position$^\dagger$ & \bo{20.88} & \bo{0.731} & \bo{0.184} \\
      Rotation           & \bt{20.23} & \bt{0.710} & \bt{0.195} \\
      Scale              & 20.22 & 0.707 & \bt{0.195} \\
      Opacity            & 20.21 & 0.709 & \bs{0.194} \\
      Position + Scale   & \bs{20.33} & \bs{0.719} & 0.199 \\
      \midrule
      \multicolumn{4}{l}{\textit{Training strategy}} \\
      \midrule
      Full Method$^\dagger$ & \bo{20.88} & \bo{0.731} & \bo{0.184} \\
      Stochastic $\rightarrow$ Mixed Loss & \bs{20.84} & \bt{0.714} & 0.212 \\
      w/o Scheduling          & \bt{20.69} & \bs{0.721} & \bs{0.191} \\
      w/o Reinitialization    & 20.58 & 0.713 & \bt{0.197} \\
      \bottomrule
    \end{tabular}
\end{table}

\subsection{Ablation Studies}

\paragraph{Perturbation Design.}
In \cref{tab:ablation_noisetype}, we evaluate different strategies for injecting perturbations into Gaussian positions. Our default configuration leverages anisotropic noise that is proportional to the scale of Gaussian axes, promoting both detail preservation and robustness. For comparison, we test isotropic variants where perturbations are scaled by the longest axis of each Gaussian, by the mean axis length, or fixed uniformly across all Gaussians at half the size threshold for determining clone/split targets. Among these, anisotropic perturbation achieves the best balance between fidelity and regularization, while isotropic variants either oversmooth fine structures or under-regularize large primitives. These results confirm the importance of accounting for the geometric shape of each Gaussian when designing perturbations. Neverthless, we find that any perturbation strategy yields better generalization than the vanilla 3DGS baseline, highlighting that introducing perturbation itself acts as a strong form of regularization, even when the design is not optimal.

\paragraph{Effect of Perturbing Different Parameters.}
To assess the sensitivity of different Gaussian attributes, we applied perturbations individually to position, scale, rotation, and opacity, as reported in \cref{tab:ablation_noiseparam}. For scale, noise was added proportional to each axis length, following our SAP strategy. Rotation was perturbed by altering each Gaussian’s orientation with small random angular offsets around the three principal axes. Opacity was perturbed by adding Gaussian noise directly to its values.

The results indicate that perturbing position is the most effective for improving generalization, consistently outperforming other parameter choices. Interestingly, applying perturbations to both position and scale does not further enhance performance; instead, its performance degraded compared to perturbing position alone. These observations suggest that inducing flatness in the position space is particularly impactful for promoting stable and generalizable geometry in 3DGS, and that geometry-based perturbation plays an important role in our framework, while other parameter groups may still be beneficial under different schedules or formulations and thus remain worth exploring in future work.

\begin{figure}[tb]
  \centering
    \includegraphics[width=\linewidth]{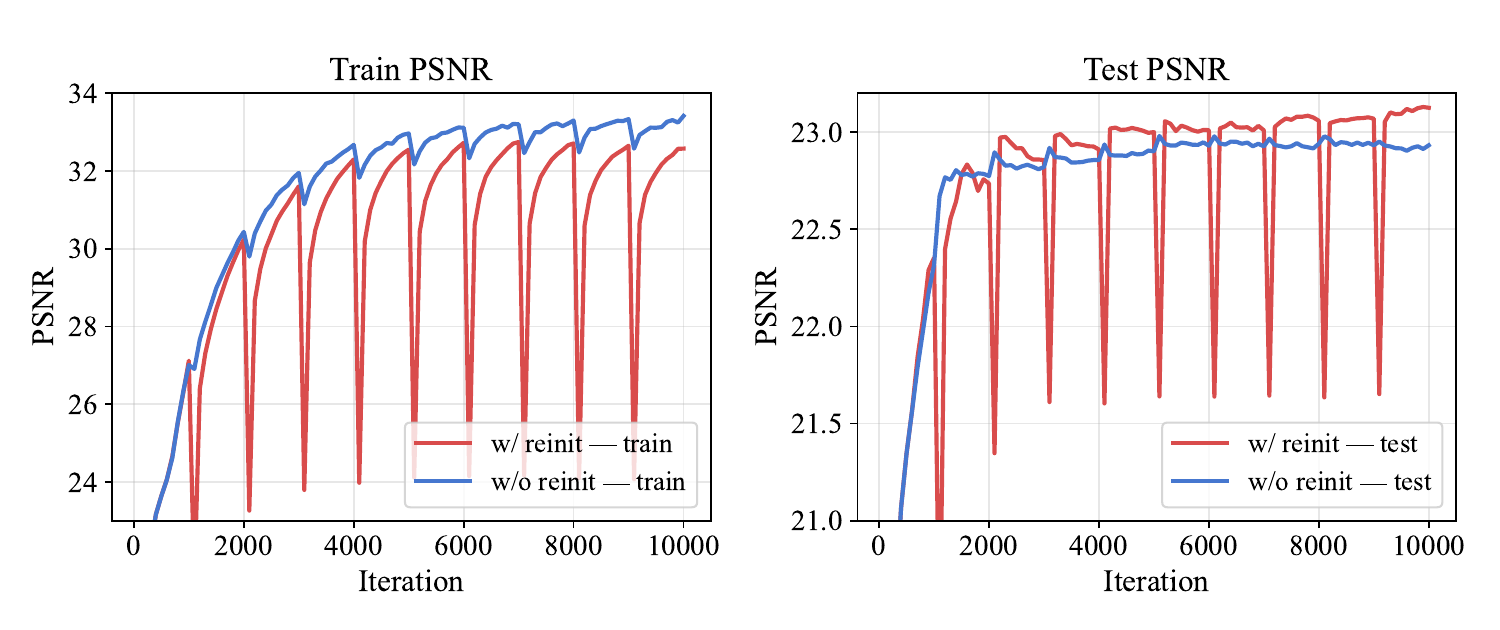}
    \caption{Training and test PSNR trajectories with and without Gaussian reinitialization on the LLFF fern scene under the 3-view setting. Without reinitialization, training PSNR keeps rising steadily while test PSNR plateaus, indicating overfitting. In contrast, periodic reinitialization moderates the growth of training PSNR but enables continuous improvements in test PSNR, suggesting better generalization. The sharper drops observed at each reset interval are due to reinitializing more diverse parameters, but performance quickly recovers and ultimately surpasses the non-reinitialized baseline. This aligns with the role of periodic reinitialization as an additional regularizer that mitigates overfitting and improves generalization.}
    \label{fig:ablation_reinit}
\end{figure}

\paragraph{Stochastic Perturbation Strategy.}
We further investigate the role of stochasticity in perturbation design. Prior flat minima studies~\cite{revisitingrwp} improve generalization by mixing losses from perturbed and unperturbed models, but this requires rendering twice per iteration. Instead, we realize a similar effect by applying perturbations stochastically during training. Concretely, each Gaussian is randomly perturbed with some probability at every iteration, which introduces variability into the optimization while keeping the training objective simple and lightweight. We found that directly relying on perturbed losses at every step tends to oversmooth high-frequency details, whereas our stochastic scheme alleviates this issue and preserves fine structures. This can also be observed in our ablation results (\cref{tab:ablation_loss_mix}).

\paragraph{Perturbation Magnitude Scheduling.}
We further validate the role of perturbation scheduling by comparing models trained with and without the progressive scaling of perturbation magnitude (\cref{tab:ablation_schedule}). Without scheduling, strong perturbations are applied from the very beginning of training, when Gaussians have not yet captured the coarse scene structure. As a result, optimization is hindered, leading to unstable convergence and degraded reconstruction quality. In contrast, our scheduling strategy gradually increases perturbation strength as training progresses, allowing Gaussians to first establish a stable representation of the scene before stronger perturbations are introduced. This results in consistently better quantitative performance, supporting our intuition that early excessive noise is detrimental, while a progressive schedule provides a more effective balance between regularization and fidelity.

\begin{table}[tb]
    \footnotesize
    \centering
    \caption{Plug-in compatibility across optimization-based, diffusion-enhanced, and feed-forward sparse-view 3DGS pipelines.}
    \label{tab:plugin}
    \renewcommand{\arraystretch}{1.0}
    \setlength{\tabcolsep}{5pt}
    \setlength{\aboverulesep}{0.35ex}
    \setlength{\belowrulesep}{0.35ex}
    \begin{adjustbox}{max width=\textwidth}
    \begin{tabular}{l l ccc}
      \toprule
      Dataset & Method & PSNR & SSIM & LPIPS \\
      \midrule
      \multicolumn{5}{l}{\textit{Optimization-based}} \\
      \midrule
      \multirow{4}{*}{LLFF~\cite{llff} 3-view}
      & FSGS~\cite{fsgs} & 20.43 & 0.682 & 0.248 \\
      & FSGS + Ours & \textbf{21.03} & \textbf{0.725} & \textbf{0.196} \\
      & DropGaussian~\cite{dropgaussian} & 20.76 & 0.713 & 0.200 \\
      & DropGaussian + Ours & \textbf{21.05} & \textbf{0.734} & \textbf{0.192} \\
      \midrule
      \multicolumn{5}{l}{\textit{Diffusion-enhanced correction}} \\
      \midrule
      \multirow{2}{*}{Mip-NeRF360~\cite{mipnerf360} 12-view}
      & Difix3D+~\cite{difix3dplus} & 19.25 & 0.554 & \textbf{0.375} \\
      & Difix3D+ + Ours & \textbf{19.63} & \textbf{0.575} & 0.386 \\
      \multirow{2}{*}{Tanks \& Temples~\cite{tanksandtemples} 12-view}
      & Difix3D+~\cite{difix3dplus} & 14.07 & 0.499 & \textbf{0.494} \\
      & Difix3D+ + Ours & \textbf{14.35} & \textbf{0.517} & 0.516 \\
      \midrule
      \multicolumn{5}{l}{\textit{Feed-forward}} \\
      \midrule
      \multirow{3}{*}{Mip-NeRF360~\cite{mipnerf360} 12-view}
      & AnySplat~\cite{anysplat} & 15.76 & 0.307 & 0.481 \\
      & AnySplat + 3DGS 3k & 16.97 & 0.338 & 0.432 \\
      & AnySplat + Ours 3k & \textbf{17.36} & \textbf{0.384} & \textbf{0.427} \\
      \multirow{3}{*}{Tanks \& Temples~\cite{tanksandtemples} 12-view}
      & AnySplat~\cite{anysplat} & 12.67 & 0.308 & 0.506 \\
      & AnySplat + 3DGS 3k & 13.58 & 0.325 & 0.469 \\
      & AnySplat + Ours 3k & \textbf{14.07} & \textbf{0.394} & \textbf{0.445} \\
      \bottomrule
    \end{tabular}
    \end{adjustbox}
\end{table}

\paragraph{Periodic Gaussian Reinitialization.}
To validate the effect of our reinitialization strategy, we conducted an ablation study on the LLFF dataset under the 3-view setting (\cref{tab:ablation_reinit}, \cref{fig:ablation_reinit}). When reinitialization is disabled, the additional periodic return-to-initialization of scale, rotation, and higher-order SH coefficients is deactivated, while opacity continues to follow the original reset mechanism introduced in~\cite{3dgs}. Under this setting, performance drops across all metrics, with PSNR decreasing by 0.3. This indicates that augmenting the standard opacity reset with periodic return-to-initialization provides a mild regularization effect during training. It serves as a complementary mechanism alongside Scale-Adaptive Perturbation, jointly mitigating overfitting and stabilizing optimization.

\subsection{Plug-in Compatibility}
Since our method acts on the optimization process rather than introducing architectural changes, it can be naturally integrated into existing sparse-view 3DGS pipelines as a plug-in module. To verify this, we apply our method to optimization-based methods (FSGS~\cite{fsgs} and DropGaussian~\cite{dropgaussian}), diffusion-based artifact correction framework (Difix3D+~\cite{difix3dplus}), and feedforward initialization followed by short post-optimization (AnySplat~\cite{anysplat}), without modifying their original pipeline designs. As shown in \cref{tab:plugin}, incorporating our optimization strategy consistently improves FSGS, DropGaussian, and AnySplat, while also improving PSNR and SSIM when combined with Difix3D+. This suggests that the flat minima regularization provided by our approach is complementary to existing techniques, and can be readily adopted to further improve their performance under sparse-view supervision.

\section{Conclusion}

We addressed the challenge of sparse-view generalization in 3D Gaussian Splatting (3DGS) from the perspective of flat minima optimization. Viewing overfitting as convergence to sharp solutions that are sensitive to positional perturbations, we proposed a lightweight training framework that adapts flat minima principles to 3DGS through scale-adaptive perturbation with stochastic application and magnitude scheduling, together with periodic reinitialization of non-positional parameters. These components regularize optimization while preserving fine details, requiring no architectural changes or external priors. Experiments on LLFF and Mip-NeRF360 datasets demonstrate improved stability and generalization under sparse-view supervision, producing sharper and more robust reconstructions. Our results also suggest that further exploring perturbation strategies for other Gaussian parameters may provide additional opportunities to improve 3DGS optimization.

\section*{Acknowledgements}
This work was supported in part by MSIT/IITP (No. RS-2022-II220680, RS-2020-II201821, RS-2019-II190421, RS-2024-00459618, RS-2024-00360227, RS-2024-00437633, RS-2024-00437102, RS-2025-25442569), MSIT/NRF (No. RS-2024-00357729), and KNPA/KIPoT (No. RS-2025-25393280).

%
%
\bibliographystyle{splncs04}
\bibliography{main}

\clearpage
\appendix
\section{Appendix}

\subsection{Additional Benchmark Results}

We evaluate our method on additional real-world outdoor, synthetic, and object-centric benchmarks, including NeO 360 dataset for outdoor settings with limited viewpoint coverage.

\begin{table}[H]
    \footnotesize
    \centering
    \caption{Additional benchmark results.}
    \label{tab:appendix_additional_benchmarks}
    \renewcommand{\arraystretch}{0.82}
    \setlength{\aboverulesep}{1pt}
    \setlength{\belowrulesep}{2.2pt}
    \setlength{\tabcolsep}{4pt}
    \begin{adjustbox}{max width=\linewidth}
    \begin{tabular}{l l ccc}
      \toprule
      \textbf{Dataset} & \textbf{Method} & \textbf{PSNR} & \textbf{SSIM} & \textbf{LPIPS} \\
      \midrule
      \multirow{3}{*}{\textit{Tanks \& Temples 12-view}}
      & \textbf{3DGS} & 13.20 & 0.457 & 0.513 \\
      & \textbf{DropGaussian} & 12.94 & 0.478 & 0.522 \\
      & \textbf{Ours} & 13.90 & 0.503 & 0.499 \\
      \midrule
      \multirow{3}{*}{\textit{Blender 8-view}}
      & \textbf{3DGS} & 21.37 & 0.851 & 0.135 \\
      & \textbf{DropGaussian} & 21.66 & 0.858 & 0.131 \\
      & \textbf{Ours} & 22.73 & 0.875 & 0.113 \\
      \midrule
      \multirow{3}{*}{\textit{DTU 3-view}}
      & \textbf{3DGS} & 18.93 & 0.824 & 0.151 \\
      & \textbf{DropGaussian} & 19.54 & 0.846 & 0.136 \\
      & \textbf{Ours} & 19.84 & 0.855 & 0.129 \\
      \midrule
      \multirow{3}{*}{\textit{NeO 360 12-view}}
      & \textbf{3DGS} & 21.73 & 0.756 & 0.319 \\
      & \textbf{DropGaussian} & 22.01 & 0.774 & 0.315 \\
      & \textbf{Ours} & 22.47 & 0.777 & 0.324 \\
      \bottomrule
    \end{tabular}
    \end{adjustbox}
\end{table}

Table~\ref{tab:appendix_additional_benchmarks} shows consistent gains over both baselines on Tanks \& Temples, Blender, and DTU; on NeO 360, our method improves PSNR and SSIM.

\subsection{Results under Increased View Counts}

We further increase the number of input views to study behavior beyond the highly sparse regime and identify when our regularization remains useful.

\begin{table}[H]
    \footnotesize
    \centering
    \caption{Results under increased view counts.}
    \label{tab:appendix_more_views}
    \renewcommand{\arraystretch}{0.82}
    \setlength{\aboverulesep}{1pt}
    \setlength{\belowrulesep}{2.2pt}
    \setlength{\tabcolsep}{5.5pt}
    \begin{adjustbox}{max width=\linewidth}
    \begin{tabular}{l l ccc}
      \toprule
      \textbf{Dataset} & \textbf{Method} & \textbf{PSNR} & \textbf{SSIM} & \textbf{LPIPS} \\
      \midrule
      \multirow{2}{*}{\textit{LLFF 15-view}}
      & \textbf{3DGS} & 27.06 & 0.887 & 0.081 \\
      & \textbf{Ours} & 27.82 & 0.904 & 0.067 \\
      \midrule
      \multirow{2}{*}{\textit{DTU 24-view}}
      & \textbf{3DGS} & 23.21 & 0.891 & 0.127 \\
      & \textbf{Ours} & 24.10 & 0.910 & 0.130 \\
      \midrule
      \multirow{2}{*}{\textit{Mip-NeRF360 24-view}}
      & \textbf{3DGS} & 22.80 & 0.708 & 0.276 \\
      & \textbf{Ours} & 24.19 & 0.771 & 0.219 \\
      \midrule
      \multirow{2}{*}{\textit{Mip-NeRF360 36-view}}
      & \textbf{3DGS} & 25.15 & 0.797 & 0.187 \\
      & \textbf{Ours} & 25.01 & 0.774 & 0.241 \\
      \bottomrule
    \end{tabular}
    \end{adjustbox}
\end{table}

Table~\ref{tab:appendix_more_views} shows gains on LLFF 15-view, DTU 24-view, and Mip-NeRF360 24-view. The gain diminishes on Mip-NeRF360 36-view, where 3DGS is better constrained.

\subsection{Robustness to SfM Quality}

To assess the robustness of our method to the quality of the SfM initialization, we degrade the initial point cloud by downsampling the SfM points at different rates. We conduct this analysis on the LLFF 3-view setting, where the reconstruction is highly sensitive to the amount and quality of the initial geometric prior.

\begin{table}[H]
    \footnotesize
    \centering
    \caption{Robustness to SfM quality.}
    \label{tab:appendix_sfm_quality}
    \renewcommand{\arraystretch}{1.0}
    \setlength{\tabcolsep}{8pt}
    \begin{adjustbox}{max width=\linewidth}
    \begin{tabular}{l ccc}
      \toprule
      \textbf{Init. SfM points} & \textbf{3DGS} & \textbf{DropGaussian} & \textbf{Ours} \\
      \midrule
      \textbf{50\% (avg: 7.1k)} & 19.55 & 20.25 & 20.58 \\
      \textbf{10\% (avg: 1.4k)} & 19.00 & 19.59 & 19.93 \\
      \textbf{5\% (avg: 0.7k)} & 18.53 & 19.13 & 19.45 \\
      \bottomrule
    \end{tabular}
    \end{adjustbox}
\end{table}

As shown in Table~\ref{tab:appendix_sfm_quality}, our method consistently outperforms both 3DGS and DropGaussian across all downsampling rates. These results indicate that our optimization remains effective even when the initial SfM point cloud becomes substantially sparse, suggesting lower sensitivity to SfM quality.

\end{document}